\documentclass[runningheads]{llncs}
\usepackage{graphicx}
\usepackage{tikz}
\usepackage{comment}
\usepackage{amsmath,amssymb} % define this before the line numbering.
\usepackage{color}
\usepackage{multirow}
\usepackage{hhline}
\usepackage{subcaption}
\usepackage{booktabs}
\usepackage{ifthen}
\usepackage{pifont}
\usepackage{breakcites}
\usepackage[title]{appendix}

\newcommand{\xmark}{\ding{55}}%
\newcommand{\mainref}[1]{{#1}}

\usepackage[accsupp]{axessibility}  % Improves PDF readability for those with disabilities.

\usepackage[pagebackref,breaklinks,colorlinks]{hyperref}

\begin{document}

\pagestyle{headings}
\mainmatter
\def\ECCVSubNumber{6479}  % Insert your submission number here

\title{Large Scale Real-World Multi-Person Tracking} % Replace with your title

% \begin{comment}
\titlerunning{ECCV-22 submission ID \ECCVSubNumber} 
\authorrunning{ECCV-22 submission ID \ECCVSubNumber} 
\author{Anonymous ECCV submission}
\institute{Paper ID \ECCVSubNumber}
% \end{comment}
%******************

% \begin{comment}
\titlerunning{Large scale Real-world Multi-Person Tracking}

\author{Bing Shuai, Alessandro Bergamo, Uta Buechler\\ Andrew Berneshawi, Alyssa Boden, Joseph Tighe}
\authorrunning{B. Shuai et al.}
\institute{
% \email{\{bshuai,bergamo,utbuech,bernea,alyboden,tighej\}@amazon.com}\\
AWS AI Labs\\
\url{https://amazon-science.github.io/tracking-dataset/personpath22.html}
}
% \end{comment}
%******************
\maketitle

\begin{abstract}
This paper presents a new large scale multi-person tracking dataset -- \texttt{PersonPath22}, which is over an order of magnitude larger than currently available high quality multi-object tracking datasets such as MOT17, HiEve, and MOT20 datasets. The lack of large scale training and test data for this task has limited the community's ability to understand the performance of their tracking systems on a wide range of scenarios and conditions such as variations in person density, actions being performed, weather, and time of day. \texttt{PersonPath22} dataset was specifically sourced to provide a wide variety of these conditions and our annotations include rich meta-data such that the performance of a tracker can be evaluated along these different dimensions. The lack of training data has also limited the ability to perform end-to-end training of tracking systems. As such, the highest performing tracking systems all rely on strong detectors trained on external image datasets. We hope that the release of this dataset will enable new lines of research that take advantage of large scale video based training data.

\keywords{Multi-object tracking, dataset, MOT}
\end{abstract}

\section{Introduction}
Large-scale datasets are the fuel that has driven the success of learning-based methods over the past decade.
The introduction of large datasets, such as ImageNet\cite{imagenet}, MSCOCO\cite{mscoco}, LSUN\cite{lsun} and Kinetics\cite{kinetics}, has enabled the development of deep learning-based models which have rapidly advanced the field of computer vision. 
Unfortunately, no such large scale dataset has been collected for multi-object tracking to date. The multi-object tracking task~\cite{sort,sort2,centertrack,siammot,fairmot,tubetk} requires detection and ID assignment of all objects for each frame in a video. In practice many current datasets have people as the only objects (multi-person), which will also be our focus.
The most popular datasets used today, MOT17~\cite{mot} and MOT20~\cite{mot20}, have just 14 and 8 videos respectively, greatly limiting the ability of researchers to develop data hungry models that require large tracking datasets as well as limiting the measure of generalizability of tracking methods given the small number of videos used for testing. 
In this work we present a new multi-person tracking dataset that is an order of magnitude larger than MOT17~\cite{mot} and MOT20~\cite{mot20}, while maintaining the high quality bar of annotation present in those datasets. 
% \joe{note: I haven't defined multi-object tracking yet, might want to have the definition early in the intro}

One reason for the lack of large scale multi-object tracking datasets is the significant cost to collecting such a dataset. The collection and annotation of these datasets is non-trivial as both the curation (sourcing) and labeling require significantly higher manual human labor than classification or detection based datasets.  
For person tracking, sourcing video is particularly challenging because though there is a large volume of video content available on the internet, it is mostly content that does not align with our target video domain or the content rights are restricted such that the videos cannot be easily included in an academic dataset. The Kinetics\cite{kinetics} dataset, for example, attempted to remove this challenge by only providing links to YouTube videos but over time those videos were removed, leaving researchers with incomplete train and test sets and making it difficult to reliably compare to other works.

In this work we collect videos from sources where we are given the rights to redistribute the content and participants have given explicit consent, such as the MEVA\cite{meva} dataset. 
% We aim to provide a large-scale multi person tracking dataset that contains diverse videos for real-world smart home or city fixed cameras. 
Our dataset consists of 236 videos captured mostly from static-mounted cameras. Approximately 80$\%$ of these videos are carefully sourced from scratch from stock footage websites and 20$\%$ are collected from existing datasets such as PathTrack\cite{pathtrack} or MEVA\cite{meva}. 
% which do not have high quality annotations \bergamo{I would remove this last part about the quality as we don't have any data to back this up}.
While building the dataset we place special importance on sourcing indoor and outdoor videos with different lightning conditions, diverse camera angles (from birds-eye view to low-angle view), varying weather conditions (sunny, raining, cloudy, night), various levels of occlusion and different crowd densities.
%\bergamo{also, we promoted the sourcing of videos where objects are either partially or fully occluded for multiple frames}.
Section \ref{sec:sourcing} presents a detailed analysis of these factors. 

In addition to sourcing, collecting high quality annotations is especially challenging for multi-object tracking datasets. This is largely due to the complexity of the task. Classification datasets~\cite{imagenet,kinetics} only require one or more labels to be tagged per entire image or video whereas detection datasets \cite{mscoco,pascal} increase the complexity by not only requiring a list of objects, but also the object's location specified by a bounding box. Multi-object tracking extends the idea of object detection even further by also requiring a unique object identifier for every labeled bounding box throughout a video recording. This annotation task is especially challenging in crowded scenes where even a human annotator could easily lose or confuse an object with another if they get partly or fully occluded. 

In this work we adopt a two stage annotation pipeline that leverages AWS SageMaker GroundTruth (an iteration of Amazon MTurk
When annotating videos for tracking, many edge cases emerge and must be handled consistently to have a meaningful measure of an algorithm's quality.
In our annotation process, we have thoroughly considered edge cases such as people with high occlusion or person reflections and defined strict protocols for dealing with each edge case. For example, we annotate reflections of people but tag such annotations specifically so they can be properly handled during training and evaluation. After carefully defining our annotation criteria, we use our trained workforce to annotate all videos from scratch.
% After annotation, we send each video for a second round of verification by at least one additional trained annotator. 
More details regarding our annotation protocol can be found in Section \ref{sec:annotation}. 

We demonstrate the benefit our large-scale dataset adds to the community by (1) comparing key statistics with existing MOT benchmarks (Section \ref{sec:dataset}) and (2) training and evaluating state-of-the-art multi-object tracking models on our dataset (Section \ref{sec:experiments}). The latter shows that our benchmark contains many challenging scenarios where current state-of-the-art models fail to perform well. 
%\uta{not sure if we can still say 'significantly under-perform' since it seems like the results are actually not too bad} \bing{I think the results would be more interesting for models that can only be trained on video dataset, such as TubeK or MeMOT.}
We hope that the publication of our dataset will drive the tracking community towards developing more robust models that can generalize to a wide variety of smart home/city scenarios.

\section{Related Work}
% We will first review existing Multi-Object Tracking (MOT) datasets with a focus on multi-person tracking. Then we review the recent progress of MOT methods, including the ones we use, to demonstrate the utility of our dataset.

\paragraph{\textbf{Multi-Object Tracking Datasets.}}
MOT is an essential part of important applications such as autonomous driving~\cite{ess2010object,rangesh2019no,rezaei2021traffic}, smart city~\cite{datta2002person,mathur2018intelligent,chandrajit2016multiple,chandrakar2022enhanced} and activity recognition~\cite{wu2007scalable,beddiar2020vision}. Especially the field of autonomous driving has grown significantly, which is also reflected in the number of large-scale benchmarks published for this scenario~\cite{kitti,cityscapes,lyft,waymo,nuscenes,argoverse,crp,yu2020bdd100k}. Some of these datasets have also been used to train and/or evaluate person tracking models \cite{quasidense,siammot}.
The challenges such benchmarks entail are fast camera motions and quick position changes of pedestrians. However, the amount of occlusions and crowdness is rather limited and thus not sufficient enough to train robust tracking models that can operate in high-occlusion scenarios.
% , i.e. a camera and/or other sensors inside or on top of a moving vehicle 
In contrast, synthetic datasets that have been specifically created for pedestrian detection/tracking in urban scenarios \cite{motsynth,jta} contain scenes with varying person densities and can therefore be very valuable for person tracking. The clear advantage is that they do not require any manual annotations. Although the quality of synthetic data improves steadily, the usage of such data is rather limited due to the apparent domain shift to real-world data.

% Synthetic data is mostly used to artificially increase the number of training samples for complex models given the limited number of real MOT videos \cite{tubetk,tokmakov2021learning}.

Recently, a few real-world MOT datasets have been proposed. For instance, CroHD ~\cite{CroHD} dataset was introduced to track pedestrain's head in crowded scenes, GMOT-40~\cite{gmot40} was proposed for the purpose of general object tracking, and 
MVMHAT~\cite{mvmhat} and MMPTRACK~\cite{mmptrack} are adopted for multi-camera multi-person tracking. In general, their sizes are a magnitude smaller than our dataset. One of the biggest real-world MOT datasets is TAO \cite{tao}, which provides a great variety of scenes. Since TAO is created for general object tracking, the number of challenging person tracking sequences is rather limited given that a large number of videos contain only a single person. Moreover, TAO provides full annotations for only a small fraction of videos, which makes it difficult to train on. 
In contrast, our dataset has been exhaustively annotated. 
% Our benchmark also contains (re-annotated) videos from existing datasets, but a great number of samples are additionally sourced from scratch from video-sharing websites. 

Finally, the MOT datasets that are most comparable to ours are HiEve \cite{hieve}, MOT17 \cite{mot} and MOT20 \cite{mot20}. HiEve consists of 32 videos (13.5\% of the size of our dataset) and provides annotations for different human-centric understanding tasks such as pedestrian tracking or pose estimation~\cite{song2021human,liu2021recent,chang2020towards}. The main goal of HiEve is to provide a set of videos that are recorded during complex events (e.g. earthquake escapes). 
\texttt{PersonPath22} dataset, on the other hand, has the objective of providing a wide variety of smart home/city scenarios during different seasons, varying lightning and weather conditions and diverse crowd densities without focusing solely on the complexity of events. 
% HiEve, for instance, does not contain any recordings during the night or challenging weather conditions. 
The most popular MOT benchmark which has also a similar purpose as ours is MOT17 \cite{mot}. The benchmark consists of 14 videos that are recorded at 9 different scenes with different lightning and camera angles. MOT20 \cite{mot} extends the MOT17 benchmark by 8 additional videos, which was specifically created for tracking in crowds. \texttt{PersonPath22} dataset also contains very crowded scenes, but provides on top of that a wide variety of pedestrian densities indoor and outdoor. 
% Moreover, we supply a significantly larger number of videos with 3 times more person tracks. 

\paragraph{\textbf{Multi-Object Tracking Methods.}} Many of the well-known MOT models follow the detection-by-tracking paradigm \cite{siammot,sort,sort2,leal2016learning,ristani2018features,wang2019exploit,xu2019spatial,xu2020train}, in which object instances are firstly detected for every frame and then they are linked across frames to form object tracks. 
% Offline trackers~\cite{zhang2008global,berclaz2011multiple,zamir2012gmcp,kim2015multiple,tang2017multiple,henschel2017improvements,ristani2018features,sheng2018heterogeneous,xu2019spatial,wang2019exploit,andriyenko2011multi,berclaz2006robust,evangelidis2008parametric,tang2017multiple} are usually time-consuming and therefore they are less likely to be applied in real-world tracking.
% formulate this task as to optimize an objective function related to a densely-connected graph in which every detected box is a node and a pair of boxes across frames are connected with an edge. Solving such a graph-based optimization problem is usually time-consuming and thus it is less likely to be applied in real-world tracking.\uta{do we need to mention offline trackers here if they are anyway no 'customers' of our dataset? For completeness it's probably good to mention, but given the space restrictions, this might be less important} 
Recently, online trackers~\cite{siammot,centernet,fairmot,quasidense,jde} have steadily gained ground by pushing the results on MOTChallenge~\cite{mot} to new highs. Those trackers are usually deep neural networks that include key models for online tracking, which include a detection model~\cite{centernet,fasterrcnn,fcos,detr}, a motion model~\cite{sort,sort2,siammot,siamfc,siamrpn++} and an optional person re-identification model~\cite{jde,fairmot}. Those models are usually jointly trained with tracking annotations, i.e. a bounding box with a unique identifier. Due to the scarcity of those annotations, self-supervised training techniques~\cite{centertrack,siammot,fairmot,pointid} were developed to leverage image-based object detection datasets for model pre-training. 
% \bing{It's good that we can claim that our dataset has value for training, but we don't have experiments to back up. So I'm refrained from saying too much about this.} 
In this work, we train and evaluate three recent state-of-the-art online trackers on our dataset.

\section{Video Sourcing}\label{sec:sourcing}

\begin{figure}[t]
    \centering
    \includegraphics[width=1.0\textwidth]{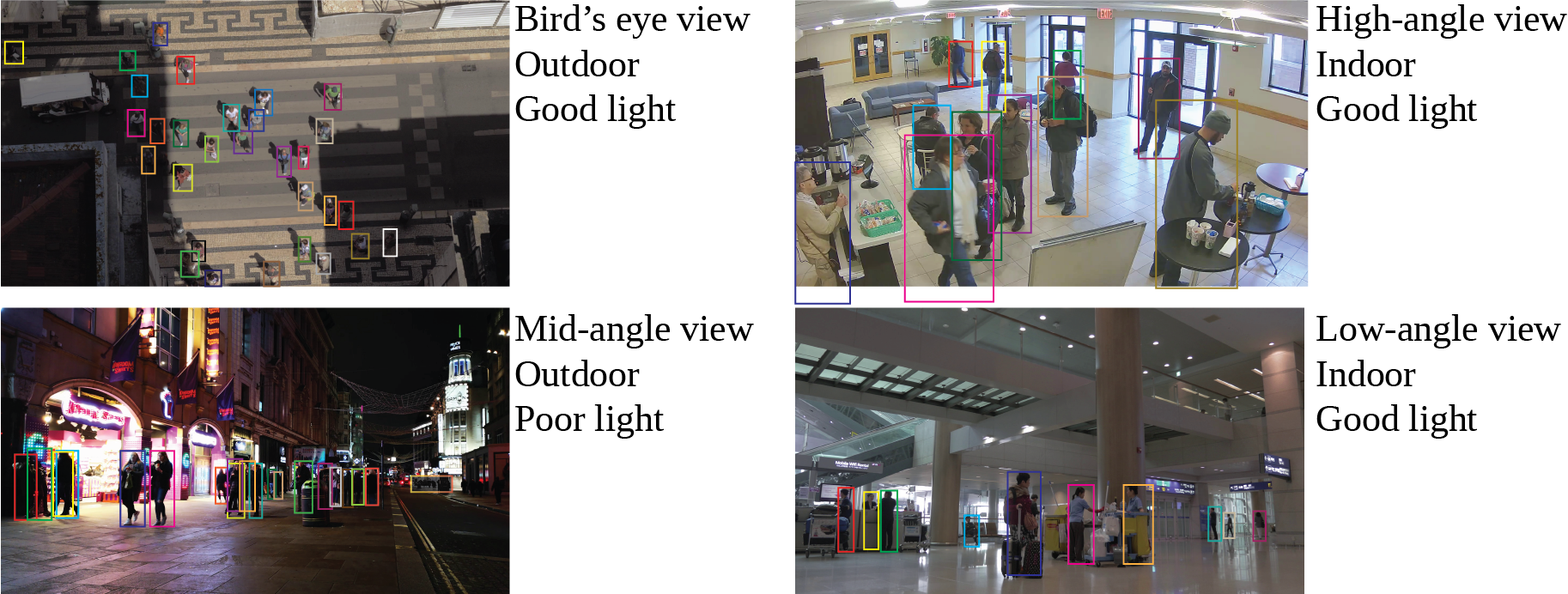}
    % \caption{ Visual examples of key frames with the corresponding annotations.
    % % \joe{this isn't referenced anywhere in the text and frame examples should have labels under them like <bird's-eye, outdoor, night, low-light> to give examples of the meta-data we will provide}
    % }
    % \begin{subfigure}[b]{0.5\textwidth}
    % \centering
    % \includegraphics[width=1.0\textwidth]{Figs/video_attributes/video_sources.png}
    % \caption{Video sources \bing{Consider moving this fig to the supplementary}}
    % \label{fig:video_source}
    % \end{subfigure}
    \caption{ \small The figure is best viewed in color.  Frames in the video are exhaustively annotated with person boxes, each of which have a unique identifier (i.e. color-coded box).  The videos in \texttt{PersonPath22} dataset cover diverse tracking scenarios in terms of  camera angles, weather / lighting condition and scenery types. 
    }
    \label{fig:video_frames}
\end{figure}

The creation of a dataset for training / evaluating person tracking algorithms needs to strike a balance between the need of (1) having videos that represent a large variety of real-world tracking applications; (2) having videos containing challenging scenarios for tracking algorithms (e.g. occlusions, small objects);
% to avoid seeing inflated evaluation metrics due to many objects that are trivial to track; 
and 
(3) ensuring that the data is collected in a responsible way such that it can be used in perpetuity.
% \footnote{For example, videos in the Kinetics~\cite{kinetics} can suddenly become unavailable as the corresponding YouTube videos are removed, hence making it impossible to precisely compare the accuracy across different works.} 
Following these guidelines, we source our dataset in two steps.

%%% DATA SOURCES SELECTION %%%
\paragraph{\textbf{Data Source Selection.}}
We select a pool of data sources based on the availability of video content suitable for tracking applications, as well as the presence of an appropriate license that allows the data to be used and remain available for academic research.
We source videos from stock video services (Fillerstock~\cite{fillerstock}, Pexels~\cite{pexels}, Pixabay~\cite{pixabay}) and from public academic datasets for human activity understanding (MEVA~\cite{meva}, Virat~\cite{virat}, PathTrack~\cite{pathtrack}) where proper licensing is available.  The breakdown of the number of videos for each data source is provided in the supplementary material.
% These different sources allow us to include videos from a large variety of domains (security/monitoring/retail), environments (indoor/outdoor/day/night), and camera types (professional security cameras but also commodity hardware). 
Note that although MEVA and Virat come with incomplete person bounding boxes annotations, we re-annotate all videos included in our dataset to ensure consistency in annotations across all data sources. We first create an initial large candidate set of videos by automatically querying content from Fillerstock, Pexels, and Pixabay using a pre-defined set of search keywords such as ``person walking in the shopping mall" (please refer to the supplementary material for the full list). The union of these videos and the videos from the public datasets form our candidate video set.

%More in detail, we downloaded \textcolor{red}{X} videos for each keyword for each source\uta{do we need to mention how much per keyword? That was quite arbitrary no?}. We also included in the candidate set all the videos from MEVA, Virat, and PathTrack.

%%% manual selection %%%
\paragraph{\textbf{Manual Selection.}}
Our initial candidate dataset includes 8000+ videos which are then manually inspected by a team of experts. The selection processes took into account the following criteria: (1) application aligned (fixed connected home or city level cameras), (2) moving crowds, (3) occlusion, (4) background variability, (5) static vs moving cameras, (6) camera position and (7) environment conditions (day/night, sunny/rain/snow/cloudy etc). More details to the mentioned criteria are elaborated in the supplementary material.
% \uta{is this too short now? I think it is enough to just mention it here and then refer to the suppl.}
In total, we select 236 videos for manual annotation and inclusion in our dataset. The cumulative temporal duration of these videos is $139$ minutes.
% Fig.~\ref{fig:video_source} reports a breakdown of the number of videos for each source.

%%%%%%%%%%%%%%%%%%%%%%%% ANNOTATION %%%%%%%%%%%%%%%%%%%%%%%%%%%%%
\section{Annotation Pipeline}\label{sec:annotation}
% We have thoroughly considered edge cases such as person reflections and defined strict protocols for dealing with each edge case.
% The following paragraphs introduce our annotation pipeline, edge cases we've considered.
% and the cost that comes with exhaustively annotating videos for person tracking. 

% \subsection{Annotation Pipeline}
% Like other multi-object tracking datasets~\cite{mot,mot20,hieve}, we exhaustively annotate all person instances with a bounding box and a unique identity. 
Annotating person boxes with identities is time-consuming and error-prone. To this end, we adopt AWS SageMaker GroundTruth (SMGT) service\footnote{\scriptsize \url{https://docs.aws.amazon.com/sagemaker/latest/dg/sms-video-object-tracking.html}} (an advanced version of Amazon Mechanical Turk). This workflow works as follows. First, the annotator draws bounding boxes for all visible people in the starting frame. In the next frame, the SMGT service leverages a pre-trained model to predict the bounding box for each annotated person.
The annotator first verify the quality of predicted bounding boxes and adjust the bounding boxes as needed. Then, the annotator draws bounding boxes for those persons that do not appear in earlier frames. 

We employ professional annotators that have been specially trained for this task. 
%Each annotator has been actively trained to perform various computer vision annotation tasks including multi-object tracking. In particular, 
We ask them to annotate every possible visible person in the video unless they are too small in size ($< 20 \times 20$ pixels) to be accurately localized or they are in a crowd. 
In the latter case, we ask the annotators to draw a bounding box with \texttt{crowd label} that includes all people in the crowd (e.g. Fig.~\ref{fig:video_frames}(bottom left)). If a person enters the area labeled as crowd, with  $>95\%$ of the person's bounding box covered by a crowd box, we label this occurrence as `ignore' to ensure that the predicted tracks are not penalized on these cases. 
%As soon as the overlap with the crowd box decreases, the track can be continued.\uta{@Alessandro: could you please check if this is what you meant?}\uta{add an example with crowd box}. 
As shown in Fig.~\ref{fig:video_frames}, we annotate with \texttt{amodal} bounding boxes, indicating that the full extent of the bounding box is annotated regardless of the visibility status of the underlying person. In addition, we also annotate the corresponding \texttt{visible} bounding boxes that only enclose the visible part of the person body. This inclusion of both annotation types give researchers the most flexibility when choosing how to train their models and evaluating these models on other datasets.

To ensure that the annotation is of high quality, we perform a second round of labeling where a separate group of annotators checks if (1) all people are annotated, (2) all bounding boxes are correctly localized and (3) the identity of a person track is consistent throughout the video. In case the annotators notice a mistake, they correct the error. 
Finally, the authors of this paper do a final verification pass on the data, sending back any videos that have errors for re-annotation. This rigorous process allows us to have high confidence of the quality of the provided annotations. 
% \joe{move later in paper to some details section} 
We first annotate at 5 frames per second. Then, we linearly interpolate those annotations and let our trained annotators verify the correctness of those interpolations for every frame and person. 

% \subsection{Person Track Categorization}
Given that not all annotated person boxes are equally interesting, and some might even be perceived as noise, we further annotate each person track with the following tags: 1), sitting/standing still person; 2),person in vehicle; 3),person on open-vehicle; 4), reflection; 5), severely occluded person; 6), person in background; 7), foreground person. These fine-grained track-level tags enable to train or evaluate models along different sets of person tracks based on the needs of various tracking applications. The definition and visual examples of those tags are provided in the supplementary materials.

\begin{figure}[t]
    \centering
    \begin{subfigure}[b]{0.3\textwidth}
    \centering
    \includegraphics[height=1.\textwidth]{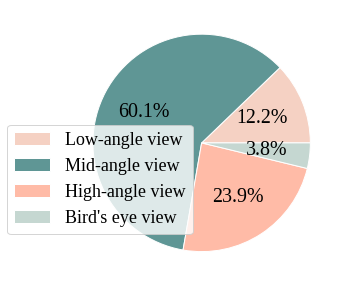}
    \caption{ Camera angles}
    \label{fig:camera_angles}
    \end{subfigure}
    \begin{subfigure}[b]{0.4\textwidth}
    \centering
    \includegraphics[height=0.75\textwidth]{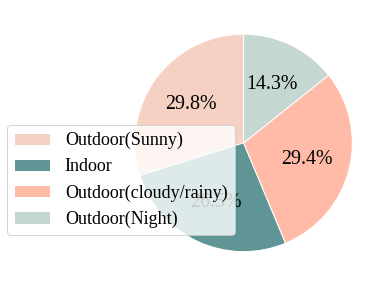}
    \caption{ Scenes / weathers}
    \label{fig:scene_weather}
    \end{subfigure}
    \begin{subfigure}[b]{0.25\textwidth}
    \centering
    \includegraphics[height=1.2\textwidth]{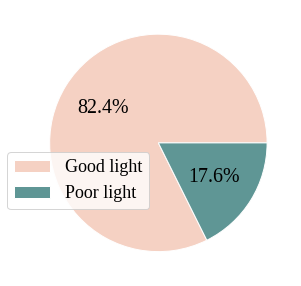}
    \caption{ Light condition}
    \label{fig:light_condition}
    \end{subfigure}
    \caption{ \small Video-level statistics of training videos in \texttt{PersonPath22} dataset.}
    \label{fig:video_statistics}
\end{figure}

\section{Dataset}\label{sec:dataset}
% Our dataset includes 236 videos that total 138.9 minutes in length. 
To understand how \texttt{PersonPath22} dataset compares to current MOT datasets we analyze various statistics of our and other publicly available datasets. We specifically compare to three popular datasets: MOT17~\cite{mot}, HiEve~\cite{hieve}, and MOT20~\cite{mot20}.

\subsection{Video-level Statistics}

\paragraph{\textbf{Camera Angles.}} We categorize the angles of the cameras that are used to record the underlying videos into four buckets: (1) bird's-eye view, (2) high-angle view, (3) mid-angle view and (4) low-angle view. Visual examples are given in Fig.~\ref{fig:video_frames}.
% When the altitude of a camera is significantly higher than / as high as / lower than the human body, the corresponding views are high-angle / mid-angle / low-angle, respectively. When the altitude of a camera is higher than a building, we consider it a bird's-eye view.\uta{not sure if we need to define exactly what we see as bird's eye view etc. We have a figure at the beginning that shows roughly how we define it. So we could remove this part from "When the altitude..."}
% Usually, in videos with bird's-eye view and high-angle view, a person is less likely to be fully occluded by other people and a person's scale does not change significantly. However, in mid/low-angle view, occlusions between people happen frequently and a person's scale varies substantially (e.g. a person is moving towards the camera), which results in challenging cases for tracking.\uta{also this paragraph is in my opinion not super essential. We talk about occlusions later and that we have them, does not really matter if they are coming from camera angles or from somewhere else, no? + in crowd density we even mention again the camera angles.}
As shown in Fig.~\ref{fig:camera_angles}, our dataset contains 143 (60.1\%) and 33 (23.9\%) videos that are recorded by mid and low-angle-view cameras, respectively. On this front, the closest dataset to ours is MOT17~\cite{mot} that has 10 (71.4\%) mid-angle-view and 4 (28.6\%) high-angle-view videos. Out of 32 videos in HiEve dataset~\cite{hieve}, only 2 (6.3\%) videos are recorded with mid-angle-view cameras and the remaining 30 (93.7\%) videos are with high-angle-view. For MOT20 dataset~\cite{mot20}, all 8 videos are captured with high-angle-view cameras.

\paragraph{\textbf{Scenes and Weather.}} We categorize the scene of a video into two buckets: (1) indoor (e.g. cafe house, mall, airport) and (2) outdoor (e.g. street, plaza, beach). For outdoor videos, we further annotate the weather condition. As shown in Fig.~\ref{fig:scene_weather}, there are 63 (26.5\%) indoor videos and the outdoor videos are evenly spread across three weather/light conditions (sunny, cloudy, night/dark). Furthermore as we show in Fig.~\ref{fig:light_condition}, there are 42 (17.6\%) videos that have poor light condition, under which tracking people becomes increasingly challenging. Overall, our dataset provides a good diversity in terms of scene types and light conditions.  In comparison, MOT17~\cite{mot} includes 2 indoor and 2 night videos.

\begin{figure}[t]
    \centering
   \begin{subfigure}[b]{0.28\textwidth}
    \centering
    \includegraphics[height=1.027\textwidth]{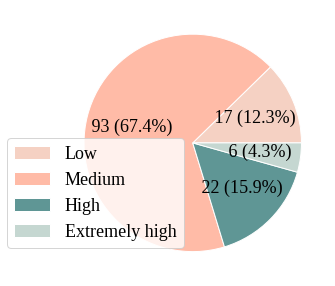}
    \caption{\scriptsize PersonPath22}
    \end{subfigure}
    \begin{subfigure}[b]{0.23\textwidth}
    \centering
    \includegraphics[height=1.25\textwidth]{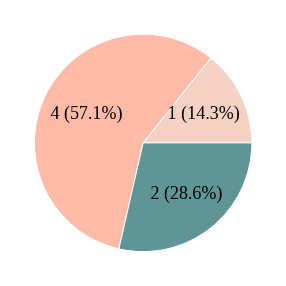}
    \caption{\scriptsize MOT17}
    \end{subfigure}
    \begin{subfigure}[b]{0.23\textwidth}
    \centering
    \includegraphics[height=1.25\textwidth]{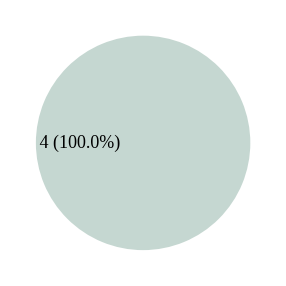}
    \caption{\scriptsize MOT20}
    \end{subfigure}
    \begin{subfigure}[b]{0.23\textwidth}
    \centering
    \includegraphics[height=1.25\textwidth]{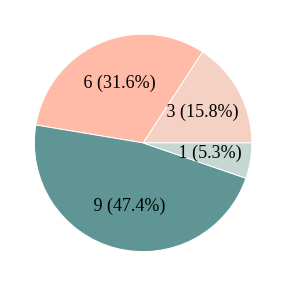}
    \caption{\scriptsize HiEve}
    \end{subfigure}
    
    \caption{\small Video-level people density distribution of training videos.}
    \label{fig:crowd_density}
\end{figure}

\paragraph{\textbf{People Density.}} We define the people density ($\mathtt{d}$) of the scene to be the average number of people per frame, based on which we categorize each video into four buckets: low density ($\mathtt{d} \leq 10$), medium density ($10 < \mathtt{d} \leq 30$), high density ($30 < \mathtt{d} \leq 60$) and extremely high density ($\mathtt{d} > 60$). As shown in Fig.~\ref{fig:crowd_density}, \texttt{PersonPath22} dataset has a similar distribution with MOT17~\cite{mot} and HiEve~\cite{hieve} dataset, although it has a significantly larger scale.
Note that although there is a positive correlation between the tracking difficulty and the people density of the video when the camera angle and scene / light condition is similar, people density is not the only indicator of difficulty level of underlying videos. For example, tracking a person in a low-angle-view video with low density can be more challenging than that in a high-density bird's-eye video due to the high level of occlusion in the low-angle-view video.

In terms of the above factors, \texttt{PersonPath22} dataset provides a set of videos that resembles a similar distribution with the current dataset MOT17~\cite{mot} but at an order of magnitude larger scale. Importantly, \texttt{PersonPath22} dataset is highly diverse which makes the training and evaluation of tracking models more representative to real-world person tracking challenges.

\subsection{Track-level Statistics} \label{sec:track_stat}

\begin{figure}[t]
\centering

\rotatebox[origin=c]{90}{\bfseries \small PersonPath22 \strut}
    \begin{subfigure}{0.26\textwidth}
        \centering
        \includegraphics[width=1.1\textwidth]{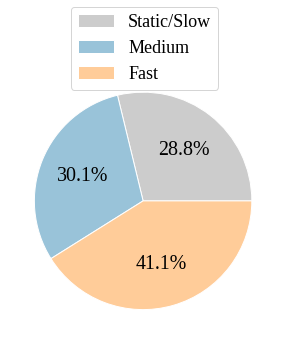}
    \end{subfigure}%
    \begin{subfigure}{0.26\textwidth}
        \centering
        \includegraphics[width=1.1\textwidth]{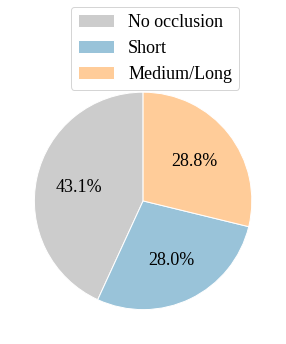}
    \end{subfigure}%
    \begin{subfigure}{0.26\textwidth}
        \centering
        \includegraphics[width=1.1\textwidth]{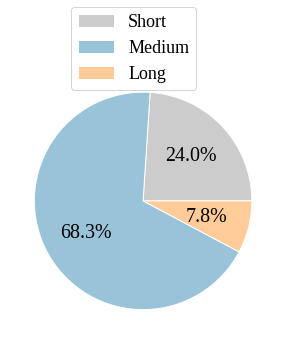}
    \end{subfigure}

\rotatebox[origin=c]{90}{\bfseries \small MOT17 \strut}
    \begin{subfigure}{0.26\textwidth}
        \centering
        \includegraphics[width=1.1\textwidth]{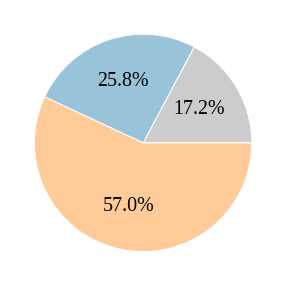}
        \caption{ \small Track speed}
        \label{fig:dataset_stat_speed}
    \end{subfigure}%
    \begin{subfigure}{0.26\textwidth}
        \centering
        \includegraphics[width=1.1\textwidth]{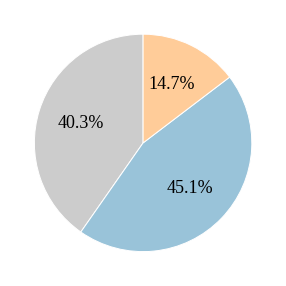}
        \caption{ \small Occlusion duration}
        \label{fig:dataset_stat_occlusion}
    \end{subfigure}%
    \begin{subfigure}{0.26\textwidth}
         \centering
         \includegraphics[width=1.1\textwidth]{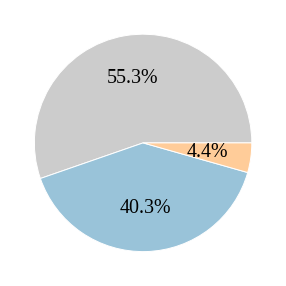}
        \caption{ \small Track duration}
        \label{fig:dataset_stat_length}
    \end{subfigure}
    \caption{ \small Key statistics of person tracks in training videos from \texttt{PersonPath22}  and \texttt{MOT17} datasets.
    % \uta{I reduced the size of the figures, which saved us quite some space, but if time allows we should create them again with a bigger font size.}
    }
\end{figure}

We further analyze the statistics of each track annotated in \texttt{PersonPath22} dataset.
We represent a person track as a temporally ordered set of bounding boxes $\mathcal{T} = [\mathtt{bb}_{t_s}, \ldots, \mathtt{bb}_{t}, \ldots, \mathtt{bb}_{t_e}]$, in which $t_s$ and $t_e$ are the start and terminal timestamp of person track $\mathcal{T}$ respectively, $\mathtt{bb}_t = (x_t, y_t, w_t, h_t)$ where $(x_t, y_t)$ is the center point coordinates of person bounding box at time $t$ and $w_t, h_t$ its width and height. 
In total, 12,150 unique person tracks are annotated, out of which 7,096 tracks are from training videos, and the remaining 5,054 from test videos. Furthermore, 7,534 tracks are labeled with ``foreground person" tag, based on which we derive the statistics of person tracks as follows.

\paragraph{\textbf{Average Track Speed.}}
We define the temporally normalized motion vector $\mathbf{m}_{(t_1 \rightarrow t_2)}$ for person track $\mathcal{T}$ between timestamp $t_1$ and $t_2$ ($t_2 > t_1$) as follows:
\begin{equation}
    \mathbf{m}_{t_1 \rightarrow t_2} = \frac{1}{\zeta \!\! \cdot \!\! (t_2 - t_1)} (x_{t_2} - x_{t_1}, y_{t_2} - y_{t_1})
\end{equation}
in which $\zeta$ is the average length 
of the person bounding box at timestamp $t_1$ and $t_2$, that is $\zeta = 0.5 * (\sqrt{(w_{t_1} \! \cdot \! h_{t_1})} + \sqrt{(w_{t_2} \! \cdot \! h_{t_2})})$. Therefore, $\mathbf{m}_{(t_1 \rightarrow t_2)}$ indicates the direction of the person's motion between timestamp $t_1$ and $t_2$, and its $\mathbf{L}_2$ norm $||\mathbf{m}_{(t_1 \rightarrow t_2)}||_2$ reflects the speed of the corresponding person within a unit of time. Then, we derive the average speed $\mathbf{v}$ for a track with the following equation
% \bergamo{If we want to save some space, we can just say "with the unbiased mean of the normalized motion vectors for a given track", and add the L2 on eq 1}
:
\begin{equation}
    \mathbf{v}_{(\mathcal{T})} = \frac{1}{|\mathbf{T}|-1}\sum_{i=2}^{|\mathbf{T}|} ||\mathbf{m}_{(\mathbf{T}[i-1] \rightarrow \mathbf{T}[i])}||_2
\end{equation}
% \joe{what is the time in seconds between i-1 and i? that will change the meaning of v} \bing{I'm not sure that I understand it. But why the meaning of v would change?}\joe{I'm basically asking the units of V. m is normalized by the difference in time ($(t_2 - t_1)$) and $\zeta$. So is it something like person lengths/second?}\bing{right.}
where $\mathbf{T} = \{t_s, \ldots, t, \ldots, t_e\}$ is a sorted list of timestamps that the person appears. We bucketize each person track to have a static/slow, medium and fast speed if $\mathbf{v}_{(\mathcal{T})} < 0.2$, $0.2 \leq \mathbf{v}_{(\mathcal{T})} < 0.6$ and $\mathbf{v}_{(\mathcal{T})} \geq 0.6$. In Fig. \ref{fig:dataset_stat_speed}, we show the distribution of track speed of our dataset in comparison with MOT17~\cite{mot}. A few videos in MOT17 are recorded with moving camera, which leads to larger portion of higher-speed person tracks (e.g. a person standing still appears to be non-static in a video with moving background). 

% In general, a static or slow-moving person is easier to track, as a simple IOU linker is sufficient to associate detected boxes across frames. However, a sophisticated motion model is essential to track fast-moving people. 

% In Fig. \ref{fig:dataset_stat_speed}(top) we show the pie chart of the bucketized track speed for all person tracks on our dataset. As shown, a large amount of person tracks are static or slow moving, however there is also a significant amount of fast-moving tracks, which makes person tracking on the dataset quite challenging.

\paragraph{\textbf{Occlusion Duration.}}
A person becomes fully occluded if its appearance feature is not discernible at that particular time. In our case,  this happens if the annotator is unable to locate their position without inferring from temporal context. Therefore, we define the occlusion duration of a person track to be the cumulative duration ($\mathbf{o}_{(\mathcal{T})}$) of the person being fully occluded. We further categorize each person track to have no, short and medium/long occlusion if $\mathbf{o}_{(\mathcal{T})} = 0$, $0 < \mathbf{o}_{(\mathcal{T})} < 2(s)$ and $\mathbf{o}_{(\mathcal{T})} \geq 2(s)$. As shown in Fig.~\ref{fig:dataset_stat_occlusion}, \texttt{PersonPath22} dataset includes a significantly higher portion of person tracks with medium/long occlusion in comparison to MOT17. In addition, there is a significant percentage (3.5\%) of person tracks whose occlusion duration is longer than 10 seconds. A particular challenge in person / object tracking is to preserve the identity consistency before and after the object becomes fully occluded. In this respect, \texttt{PersonPath22} dataset provides challenging and interesting cases. 

% For each person track, In Fig.~\ref{fig:dataset_stat_occlusion}, we show the pie chart of the duration of tracks while being not / shortly / long occluded. A large portion of person tracks are occluded by at least 1 second, which is usually the limit of existing tracking models. There are also a significant amount of long-occluded person tracks (longer than 10 seconds) that are particularly challenging for a model to keep their identities consistent.

\paragraph{\textbf{Track Duration.}} The duration of a track ($\mathbf{\l}_{(\mathcal{T})}$) is defined as the time range between the first and last appearance of the person in the video, that is  $\mathbf{\l}_{(\mathcal{T})} = t_e - t_s$. We classify each person track to be short, medium and long if $\mathbf{\l}_{(\mathcal{T})} < 5(s)$, $5 \leq \mathbf{\l}_{(\mathcal{T})} < 30(s)$ and $\mathbf{\l}_{(\mathcal{T})} \geq 30(s)$ respectively. As shown in Fig.~\ref{fig:dataset_stat_length}, person tracks in our dataset tend to be longer in contrast to MOT17~\cite{mot}. Considering that tracking a person in a longer duration is both interesting and technically challenging, \texttt{PersonPath22} dataset offers valuable testing cases along this aspect.

% . In this case, maintaining a consistent identity for those person tracks becomes nontrivial. As shown in Fig.~\ref{fig:dataset_stat_length}, there are a significant amount of person tracks that exceed 30 seconds, beyond which we categorize a person track to be long-duration.\uta{we might need to mention here what long/medium/short means in seconds}

\begin{table}[t]
    \centering
    \begin{tabular}{lcccccc}
    \toprule
         \multirow{2}{*}{Dataset} & \#Videos & Length & \#Annotated & \#Person & Min & Min.  \\
         &  & (secs) & Frames  & Tracks & Res. & FPS \\
         \midrule
        %  KITTI$^*$\cite{kitti} & 21[50] & 800  & 8,008 & 13,408 & 204 & 1224x370 & 10 & 10\cite{tao} \\
        %  CRP$^*$\cite{crp} & 7[7] & 8,722  & 20,999 & 32,351 & 4,239 & 1280x720 & 30 & N/A \\
        %  TAO$^+$\cite{tao} & 1,244[2,907] & 44,340  & 38,487 & 86,681 & 4,058 & 640x480 & 8 & 1 \\
        %  \midrule
         HiEve\cite{hieve} & 19 & 1,842 & 32,929  & 1,736 & 352x258  & 15 \\
         MOT17\cite{mot} & 7 & 215  & 5,316   & 546 & 640x480 & 14 \\
         MOT20\cite{mot20} & 4 & 357  & 8,931   & 2,215 & 1173x880 & 25\\
         PersonPath22 & 138 & 4,736 &  118,685 & 7,096 & 720x480& 15 \\
    \bottomrule
    \end{tabular}
    \caption{ \small Comparison of dataset statistics (of training set) between \texttt{PersonPath22} and existing datasets. Annotated Frames refer to the frames that are manually annotated and those that are automatically interpolated and then manually verified.
    % (+val if available) 
    % set given that we do not have access to ground-truth annotations of the test set for all the datasets. 
    % The number of scenes (3rd column) has been counted manually if possible. 
    % The abbreviation \texttt{Ann.} stands for \texttt{Annotation}.
    % \bing{Do we need numbers from KITTI,CRP and TAO here? Need votes here.}\uta{I would leave TAO, but kitty and CRP can be IMO removed}
    % \bing{Let's vote which number to keep? (Ours) or (Our Interp) or Both!}\uta{I vote for Ours Interp}
    % \bing{Probably we also need to add BDD dataset.}\uta{see my comments in Slack. I think including BDD would make look our dataset much worse since they have 130.6k tracks and 3M bounding boxes. We could also remove the autonomous driving datasets from this table as we anyway said in related work, that these kind of videos are not good enough for training considering the scenarios we want to cover here.}
    }
    
    \label{tab:comparison}
\end{table}

In Tab.~\ref{tab:comparison}, we further compare \texttt{PersonPath22} with existing person tracking datasets. In comparison to MOT17~\cite{mot}, the most popular dataset for multi-person tracking research, \texttt{PersonPath22} dataset includes an order of magnitude larger number of unique person tracks and videos. 
% This larger scale of dataset would potentially unleash the power of end-to-end multi-object tracking model such as TubeK~\cite{tubetk} and MeMOT\uta{not sure if it makes sense to mention them here since we don't show any results on it in the experimental section}.
Although MOT20~\cite{mot20} includes more annotated person tracks, their scope is specifically for tracking people in crowds.
% \uta{why does MOT20 contain more tracks? they contain 2,332 and we have 6,933} \bing{Because they have thousands of tracks for some crazy videos.}
Both the diversity of videos and the person tracks in our dataset are unparalleled w.r.t other dataset including HiEve~\cite{hieve}, which makes it a more challenging and realistic evaluation benchmark for multi-person tracking.

\subsection{Benchmarking} \label{sec:benchmark}
We randomly split the videos with 60\% train and 40\% test. 
% \joe{did we make sure the same scene isn't in both test and train?} \bing{We had a discussion on Friday, and the consensus is that we don't not change our existing split function due to time constraint. In current split, two videos that have the same scene can appear in the same split.} \joe{ok lets fix for camera ready} 
To make sure that both subsets follow a similar distribution, we perform the split for each video source separately. Overall, there are 138 train and 98 test videos, and we treat it as the official split of this dataset. The statistics for both splits are listed in the supplementary materials.

%In order to provide annotation for every frame, we adopt the temporal interpolation algorithm to generate the annotation for non-keyframes. We further ask the annotators to verify the correctness of those annotation and rectify them (i.e. adjust the bounding box) in case of errors, which rarely happens. \bing{TODO, Move this earlier sentences to annotation section.}\joe{I think it is in the earlier section now removing}  

We only evaluate on keyframes for bounding boxes with the ``foreground person" and ``standing / sitting still person" tag that aren't fully occluded. The key frames are identical with those used for manual annotation, so we are evaluating the results at 5FPS. With this evaluation protocol, we are discounting the influence from the detection failures but implicitly amplifying the effect from identity inconsistency. By doing this the missed detection on a fully occluded person are not penalized. We argue that it's more important to keep the identity prediction consistent before and after the person is fully occluded rather than inferring bounding boxes for a person that is not visible.

\section{Experiments}\label{sec:experiments}
% \bing{Finish passing until here, will start from here later. }

We evaluate three recent state-of-the-art online trackers on \texttt{PersonPath22} dataset. Each tracker includes a person detection and person identity association model, which are introduced as below.

\textit{CenterTrack~\cite{centertrack}} is a single-stage online tracking model that performs joint detection and tracking and is built upon the CenterNet~\cite{centernet} framework. The model takes as input (1) the previous RGB frame, (2) the current RGB frame, and (3) a heatmap with the tracked object centers. The model predicts the object boxes for the current frame, conditioned on the tracking center points that are provided as input. In addition, the model outputs the estimated offset motion vectors, based on which an online solver is used to link the boxes across frames.

\textit{SiamMOT~\cite{siammot}} is a two-stage tracking model which uses Faster R-CNN~\cite{fasterrcnn} for its person detection model. %Similar to CenterTrack~\cite{centertrack}, 
A Siamese-based tracker~\cite{held2016learning,siamcar} is incorporated in the network as a motion model to associate the detection bounding boxes across frames. In this work, we use the best-performing motion model, EMM, as suggested in the original paper.

\textit{FairMOT~\cite{fairmot}} is a single-stage tracking model that uses CenterNet~\cite{centernet} as person detection model. In addition to CenterNet, this method adopts a parallel branch to extract a feature vector (embedding) for each person instance. Finally, the affinity between the person's location and its embedding, together with a motion model (Kalman filter) are used to link detected people across frames.

We choose the above three models as they cover both single-stage and two-stage detection models. Besides, they cover two mainstream linking techniques: CenterTrack~\cite{centertrack} and SiamMOT~\cite{siammot} use learned motion models for bounding box linking,  whereas FairMOT~\cite{fairmot} leverages the similarity of person embeddings.  

\paragraph{\textbf{Implementation details.}} All models use DLA-34~\cite{dla} as feature backbone, and they are pre-trained on the CrowdHuman dataset~\cite{shao2018crowdhuman}. We use the official open-source implementations for all the algorithms provided by the original authors. 
% For all methods, we select the best set of hyper-parameters by using a grid search to maximize the MOTA and IDF1 metrics on the test set. \joe{you can't maximize both, what tradeoff did we make?} 
We train and evaluate the model with \texttt{amodal} bounding boxes. Please refer to the supplementary materials for more details. 

\paragraph{\textbf{Evaluation metrics.}} Following other literature, we report standard tracking metrics including MOTA and IDF1. In general, MOTA measures the overall performance of the end-to-end tracking system by accounting for both the detection and data association performance. IDF1, on the other hand, specifically indicates the performance of predicted identity consistency. For more details on these metrics we refer the reader to~\cite{mota,idf1}. 
% \joe{what happened to HOTA? I strongly think we should have HOTA metrics!} \bing{Due to time constrain, it's not realistic to report them in the main paper. HOTA evaluation codes haven't been integrated in our code base. We can however report them in the supplementary materials.}

% \paragraph{Test Data.} As mentioned in Section \ref{sec:annotation}, we first annotate with 5 FPS and linearly interpolate between those annotations to receive labels for every frame (which are then verified by annotators). When evaluating on our dataset, we only utilize the test frames that are not interpolated (i.e. we evaluate on 5FPS) because.... \uta{@Bing could you add a reason after the 'because' part? I think you know better what to write there}\joe{yeah all this caveat makes it seem we only did the interpolated to make the numbers look better, if we aren't using the annotations why do we even include them?}\bing{The interpolation part is to make sure that the annotation follows the existing protocol, as some methods may need dense annotation for training. However, we don't think this is the right way to do during evaluation. I explained in 5.3 why we only evaluate on key frames.}

%%%%%%%%%%%%%%%%  MODEL EVALUATIONS  %%%%%%%%%%%%%%%%%%%%%%%%%%%%%%%%%%%%%%%%%%%%%%%%
\subsection{Model Evaluation}
% NOTE: CenterTrack results: https://quip-amazon.com/hO4tAECMhovF/CenterTrack-baseline
\begin{table}[t]
    \centering
    \small
    \begin{tabular}{lcccccc}
        \toprule
        Methods & Occlusion Filter \ \ & IDF1 ($\uparrow$) & MOTA ($\uparrow$) & FP ($\downarrow$) & FN ($\downarrow$) & IDsw ($\downarrow$)  \\
        \midrule
        % paper draft:
        % CenterTrack~\cite{centertrack} & \xmark & 43.3 & 53.2 & 24780 & 99978 & 10774 \\
        % SiamMOT~\cite{siammot} & \xmark &  50.3 & 60.5 & 13625 & 91691 &  9125\\
        % FairMOT~\cite{fairmot} & \xmark &  57.5 & 55.7 & 14872 & 108547 & 5182\\
        
        CenterTrack~\cite{centertrack} & \xmark & 43.04  & 52.31  & 24611 & 107037 & 10487 \\
        SiamMOT~\cite{siammot} & \xmark &  49.84  & 59.56  & 13268 & 98069 & 9201 \\
        FairMOT~\cite{fairmot} & \xmark &  56.52  & 54.29  & 14568 & 116495 & 5179 \\
        
        \midrule
        % paper draft:
        % CenterTrack~\cite{centertrack} & \checkmark &  46.7 & 60.3 & 24540 & 65415 & 10601 \\
        % SiamMOT~\cite{siammot} & \checkmark &  54.2 & 68.6 & 13570 & 57104 & 8868\\
        % FairMOT~\cite{fairmot} & \checkmark &  62.0 & 63.4 & 14843 & 72986 &  5092\\

        CenterTrack~\cite{centertrack} & \checkmark &  46.36  & 59.28  & 24340 & 71550 & 10319 \\
        SiamMOT~\cite{siammot} & \checkmark &  53.71  & 67.52  & 13217 & 62543 & 8942 \\
        FairMOT~\cite{fairmot} & \checkmark &  61.05  & 61.79  & 14540 & 80034 & 5095 \\
        \midrule
        IdFree~\cite{pointid}  & \checkmark &  63.1   & 68.6  & 9218  & 66573 & 6148 \\
        TrackFormer~\cite{trackformer} & \checkmark &  57.1 & 69.7 & 23138 & 47303 & 8633 \\
        ByteTrack~\cite{bytetrack} & \checkmark & 66.8 & 75.4 & 17214 & 40902 & 5931 \\
        \bottomrule
    \end{tabular}
    
    \caption{\small Result comparison on the test split of \texttt{PersonPath22} dataset. Occlusion filter means that only bounding boxes without being tagged as fully occluded are used during evaluation.
    % \bing{I identify another issue during our filtering logic that can undercount FP when I try to make sense of the results here, CR has been sent out. Results in the following table needs to be re-generated based on the new eval codes.}\bergamo{I re-ran the evals and updated the tables}
    }
    \label{tab:results_baselines}
\end{table}

In Tab.~\ref{tab:results_baselines}, we show the results of three recent online trackers. In the default evaluation protocol~\cite{mot,mot20,hieve}, all valid person boxes on key frames are evaluated. Under such setting, all models achieve relatively low MOTA and IDF1 in comparison to the performance on MOT17~\cite{mot} and HiEve~\cite{hieve}, which underscores the challenges of our dataset. As expected, the detection failure (False Positive (FP) and False Negative (FN)) heavily influences the MOTA metric. We observe that a significant number of detection failures results from missed detections when a person becomes fully occluded. As we elaborated in Sec.~\ref{sec:benchmark}, we should not heavily penalize those missed detections as long as the predicted identity is consistent before and after the occlusion happens. To this end, we apply an occlusion filter process to exclude those boxes tagged as being fully occluded from evaluation. As shown in Tab.~\ref{tab:results_baselines}, FN is significantly decreased, which lifts MOTA by a large margin.  
Additionally, after applying the filter, a person track with occlusion is ``reduced" in length, which in-turn benefits IDF1. Nonetheless, the improvement of IDF1 is less significant than that of MOTA.

As shown in Tab.~\ref{tab:results_baselines}, SiamMOT achieves significantly higher MOTA compared to CenterTrack and FairMOT. We conjecture that its underlying detector -- Faster-RCNN~\cite{fasterrcnn} -- works better than CenterNet~\cite{centernet} which underlies the other two tracking models. To validate it, we run inference of the two underlying detectors — FRCNN~\cite{fasterrcnn} and CenterNet~\cite{centernet} on the test set. FRCNN achieves 82.03\%
AP@0.5 and CenterNet achieves 78.51\% AP@0.5.\footnote{We encourage the researchers report detection AP@0.5 of their tracking models on our dataset.} Not surprisingly, FairMOT achieves a significantly better IDF1 than the other two motion-based tracking models, despite the fact that the detected boxes have more errors than that of SiamMOT. This result suggests that person re-identification is essential for tracking models to preserve the identity consistency of predicted tracks in the case of occlusion.

We also evaluated a few recent state-of-the-art models on the PersonPath22 dataset, including 1), zero-shot IdFree~\cite{pointid} model in which the embedding component is trained without any person identity annotation, 2), TrackFormer~\cite{trackformer} whose underlying detector is transformer based Detr~\cite{detr}, as well as 3), ByteTrack~\cite{bytetrack} that is based on the state-of-the-art singe-tage YOLOX detector~\cite{yolox}. As can be clearly seen, ByteTrack achieves the best MOTA and IDF1, and the zero-shot IDFree model outperforms most recent state-of-the-art tracking models even though it is not trained on the target PersonPath22 dataset. We encourage the researchers to compare their models on both settings if applicable.

% Tab.~\ref{tab:results_baselines} shows the results of recent end-to-end online trackers trained either on the CrowdHuman dataset~\cite{shao2018crowdhuman} alone (top half of the table), or trained on both CrowdHuman and our dataset (bottom half). 
% %In both scenarios, the hyper-parameters of all the models were separately tuned by using a grid search to maximize the MOTA and IDF1 metrics on the test set\uta{that's already mentioned in implementation details above, would remove it here}. For more details regarding the training and fine-tuning procedures, please refer to the supplementary materials.
% We performed an in-depth analysis to study how the above state-of-the-art models perform under different conditions.

\begin{table}[t]
\begin{minipage}[t]{0.45\textwidth}
    \small
    \centering
    \begin{tabular}{lcccc}
        \toprule
        \multirow{2}{*}{Method}  & \multicolumn{2}{c}{IDF1($\uparrow$)} & \multicolumn{2}{c}{MOTA($\uparrow$)}  \\
                     & small & large & small & large \\
        \midrule
        % paper draft submission:
        % CenterTrack  & 34.7 & 52.5 & 36.2 & 69.6 \\
        % SiamMOT  & 48.0 & 56.5 & 51.9 & 74.8 \\
        % FairMOT  & 52.8 & 66.5 & 44.8 & 71.8 \\
        
        CenterTrack  & 34.3  & 52.7  & 34.5  & 70.1 \\
        SiamMOT  & 47.1  & 56.6  & 49.9  & 75.2 \\
        FairMOT  & 50.2  & 66.6  & 40.8  & 72.0 \\
        \bottomrule
    \end{tabular}
    \subcaption{\small Results for tracks associated to small-size vs large-size persons. 
    % A \textit{small} size person tracklet is categorized as such if the average areas of the associated bounding boxes is smaller than $1\%$ relative to the video frame area.
    }\label{tab:small_large_size}
\end{minipage}
\hfill
\begin{minipage}[t]{0.5\textwidth}
    \small
    \centering
    \begin{tabular}{lcccc}
        \toprule
        \multirow{2}{*}{Method}  & \multicolumn{2}{c}{IDF1($\uparrow$)} & \multicolumn{2}{c}{MOTA($\uparrow$)}  \\
                 & static & moving & static & moving \\
        \midrule
        % paper draft submission:
        % CenterTrack  & 45.9 & 43.0 & 58.6 & 44.1 \\
        % SiamMOT  & 53.2 & 52.5 & 69.4 & 57.6 \\
        % FairMOT  & 60.7 & 60.4 & 62.7 & 53.7 \\
        
        CenterTrack  & 45.3  & 43.2  & 57.1  & 43.7 \\
        SiamMOT  & 52.6  & 52.6  & 67.2  & 57.9 \\
        FairMOT  & 59.3  & 60.2  & 60.0  & 53.7 \\
        \bottomrule
    \end{tabular}
    \subcaption{\small Results for static-to-slow vs medium-to-fast moving tracks. 
    % \bing{The numbers for CenterTrack also don't make sense.}\bergamo{I think it makes sense, MOTA is higher for static objects, i.e. tracking static objects is easier. The same trend is for all the algorithms}
    }
    \label{tab:static_moving}
\end{minipage}
\begin{minipage}[t]{0.45\textwidth}
    \small
    \centering
    \begin{tabular}{lcccc}
        \toprule
        \multirow{2}{*}{Method}  & \multicolumn{2}{c}{IDF1($\uparrow$)} & \multicolumn{2}{c}{MOTA($\uparrow$)}  \\
                     & long & short & long & short \\
        \midrule
        % paper draft submission:
        % CenterTrack  & 33.0 & 51.0 & 39.0 & 59.2 \\
        % SiamMOT  & 41.0 & 59.3 & 53.9 & 69.8 \\
        % FairMOT  & 46.5 & 68.4 & 49.4 & 64.4\\
        
        CenterTrack  & 32.5  & 51.2  & 37.6  & 59.6 \\
        SiamMOT  & 39.8  & 59.6  & 51.4  & 70.3 \\
        FairMOT  & 44.3  & 68.3  & 45.2  & 64.4 \\
        \bottomrule
    \end{tabular}
    \subcaption{\small Results for tracks with medium-to-long vs short-to-no occlusions.}
    % An occlusion is defined as \textit{long} if it lasts for at least 2 seconds. 
    % \bing{The numbers for CenterTrack don't make sense especially MOTA.}\bergamo{I fixed the bug where the occlusion/interpolate label was not considered, and the trend is now the same for all the algorithms. I took a look at some videos, and I think the numbers make sense: for the tracks categorized as "long", the subject is never either fully occluded or not-occluded but it is often partially occluded (and quite a lot sometimes), which makes the tracking more difficult; moreover, the "short" occluded tracks also contain tracks with no occlusion at all, which are of course easier.  also I found several GT issues for tracks that have the occluded/interpolate tag, where for many frames the bbox does not have neither the occluded nor the interpolate tag (e.g. $uid\_vid\_00235.mp4$ $entity\_id 8$). IMO we could drop this figure, but let's discuss tomorrow morning}}
    \label{tab:long_occluded}
\end{minipage}
\hfill
\begin{minipage}[t]{0.5\textwidth}
    \small
    \centering
    \begin{tabular}{lcccccc}
        \toprule
        \multirow{2}{*}{Method}  & \multicolumn{3}{c}{IDF1($\uparrow$)} & \multicolumn{3}{c}{MOTA($\uparrow$)}  \\
                 & \texttt{s} & \texttt{m} & \texttt{l} & \texttt{s} & \texttt{m} & \texttt{l} \\
        \midrule
        % paper draft submission:
        % CenterTrack  & 44.1 & 47.3 & 41.9 &    -1.8    & 52.4 & 63.7 \\
        % SiamMOT  & 51.9 & 54.6 & 51.2 &        15.9 &  62.5 & 75.2 \\
        % FairMOT  & 53.0 & 62.1 & 60.6 &        14.5 & 58.7 & 67.7 \\
        
        CenterTrack  & 42.9  & 47.4  & 41.4  & -3.9  & 51.9  & 62.2 \\
        SiamMOT  & 51.3  & 54.4  & 50.7  & 16.4  & 62.1  & 73.0 \\
        FairMOT  & 52.3  & 61.7  & 58.9  & 13.2  & 58.2  & 64.6 \\
        \bottomrule
    \end{tabular}
    \subcaption{\small Results for tracks with short (\texttt{s}), medium (\texttt{m}) and long (\texttt{l}) duration.
    % \bing{Why performance for tracks with short duration have such bad MOTA.}
    % \textit{short} is less than 5 seconds, and \textit{long} is above 30 seconds.
    }
    \label{tab:track_duration}
\end{minipage}
\caption{\small Result comparison of models on different subsets of person tracks.}
\end{table}
\subsection{In-Depth Model Analysis}
\paragraph{\textbf{Small-Size Person Tracking.}}
Being able to correctly track small scale objects is important for real-world application scenarios. We categorize a person track as \textit{small} if the average areas of the associated bounding boxes is smaller than $0.5\%$ relative to the video frame area. 
% \bing{This size may not be small. In COCO, objects are categorized as small if there size is smaller than $32 \times 32$, right? So I think your initial 0.5\% threshold makes more sense.}\bergamo{I changed it back to 0.5\% and updated the tables}
% total_numer_or_tracks: 5123
% total number of tracks before filtering: 3631
% num_tracklets_total for objects: small: 1624
% num_tracklets_total for objects: large: 2007
% \bing{@Alessandro, how many tracks are categorized into small/large bucket? Probably we should put it here.}\bergamo{I added the text}
For example, any bounding box whose area is smaller than $50 \times 90$ for a standard 720p video is considered small in size. In our test set, 1,624 tracks are categorized as ``small".
% and 2,007 tracks are categorized as ``large".
As shown in Tab.~\ref{tab:small_large_size}, there is a significant performance gap between tracking large-size and small-size persons on both MOTA and IDF1. This is expected as both detecting and re-identifying low-resolution objects remains a major challenge. 

% yet it still represents one of the major failure points for modern state-of-the-art trackers. This is due to the fact that the size of the person can vary from being as small as 0.1\% relative to frame size up to 50\%. In Tab.~\ref{tab:small_large_size} we quantitatively study this issue, where we observed that both MOTA and IDF1 drops by more than 15 points when tracking small objects. \joe{anything else to say?}

\paragraph{\textbf{Static vs. Moving.}}
In real-world scenarios, video sequences contain a mix of static and moving objects. For example, people might be sitting on chairs or benches (e.g. at a park or in a waiting room), as well as standing and not moving (e.g. waiting for the pedestrian green light). We find that the presence of such objects can inflate the evaluation metrics given the fact that tracking static objects is perceptually easier than tracking moving ones. This is because static objects do not require sophisticated motion models and do not exhibit any change in appearance over time unless they are occluded. In Tab.~\ref{tab:static_moving}, we show the performance for static vs. moving objects. Overall, MOTA is significantly higher for static tracks than for moving tracks, which indicates that static/slow-moving people are easier to be detected in our dataset.  However, IDF1 performance is similar for both set of tracks, which suggests that the person's motion velocity is not strongly correlated with of its level of tracking difficulty level in our dataset.

% all evaluated models can link the detected person equally well on our dataset regardless its motion velocity.

\paragraph{\textbf{Tracks with Full Occlusion.}}
% We define a person track to be \textit{fully occluded } if the given person is fully occluded for more than 1 second. \joe{isn't this a different definition than what we had above?} 
Being able to track such scenarios is of great importance in real-world tracking applications, especially when the camera is close to the ground where person-to-person occlusion is common. Tracking through full occlusion and keeping its identity unchanged is challenging in particular in video sequences where a large number of people are present. 
% This is due to the fact that it increases the chances of having two or more people with similar visual appearances.
To this end, we report results on tracks with short-to-no occlusion and with medium-to-long occlusion, which are defined in Sec.~\ref{sec:track_stat}. As shown in Tab.~\ref{tab:long_occluded}, both the MOTA and IDF1 are substantially lower for tracks with medium-to-long occlusion. In this case, people are more likely to be partially occluded, which leads to more detection failures that contributes to lower MOTA. 
The huge gap in terms of IDF1 for all models suggests that preserving the same identity before and after the person is occluded is challenging and we hope that future research can improve performance for online trackers to track through long occlusion.

% For these reasons, we here provide a break-down in terms of how current state-of-the-art trackers perform in such conditions. Tab.~\ref{tab:long_occluded} shows the results of our analysis.

\paragraph{\textbf{Track Duration.}} 
% We define a person track to be \textit{long} if its duration is greater than 30 seconds.
In Tab.~\ref{tab:track_duration}, we show the break-down results for tracks with short, medium and long duration, as defined in Sec.~\ref{sec:track_stat}. There are a few interesting observations: 1) the IDF1 for long-duration tracks is the lowest, despite the fact that its corresponding MOTA is the highest. We find that this happens because long-duration tracks usually appear in high-angle view cameras (e.g. MEVA~\cite{meva}, Virat~\cite{virat}) in \texttt{PersonPath22} dataset, therefore detecting person in those videos is easier, which positively correlates with a higher MOTA; 2) the MOTA for short-duration track is abysmal, although it has a decent IDF1. We notice that the presence of short tracks are correlated with various challenging occlusion scenarios, for example, short tracks are associated to people in large crowds or people walking behind various objects (trees, vehicles), where the people are first visible, then become partially-occluded and disappear quickly.
The challenges presented in short, medium, and long tracks are diverse and depending on the application each could be important. Thus we hope that researchers will adopt the practice of reporting metrics on these three categories separately in the future to give further insight into their model performance.
%In the future, we hope that more research can be directed to improving the robustness for tracking objects for a longer period of time, which is of great interest in real-world applications.
% Following long tracks is of great interest in real-world applications, yet it is far from solved in the tracking community.
% This is because preserving the identity for long tracks is very challenging as the person can go through large appearance changes and different occlusion states in a longer temporal horizon. 

In summary, \texttt{PersonPath22} dataset provides interesting and challenging cases for real-world tracking that includes various duration tracks, tracks with medium-to-long occlusion and small-size person tracks, on which existing state-of-the-art online trackers struggle. 
% We hope that our dataset would spark more interesting research to tackle those challenges in real-world tracking scenarios.\uta{we mentioned that now very often in the paper, also at the end of conclusion, would remove it here (previous sentence)}

% \joe{all the topics in the section feel a bit under whelming. What do we want to say about these results? I feel like we need to make some interesting observations. Lets brainstorm some here and we can craft the key ones into the paper}\bing{The idea is to call out the tracking challenging that the existing models fails and hopefully guide other researchers to focus on this in future research. I add a few points to each paragraph, and more detailed description is needed when all results are filled in.}

%Robust data association is essential to produce a consistent identity for the above person tracks. However, their effects to the evaluation metric can be averaged out by the large number of short-duration tracks. In particular, we pay special attention to the metric that reflects the identity consistency of the predicted tracks, such as IDF1. Overall, there are XXX person tracks that satisfy the above conditions.

% * Models trained for MOT evaluated on this dataset
%     * FairMOT
%     * Center Track
%     * SiamMOT
%     * ReID Based Network
    
% * Model trained on this dataset evaluated on MOT17 (train)
%     * FairMOT
%     * CenterTrack
%     * SiamMOT
%     * ReID Based Network

\section{Conclusion and Discussion}
In this paper, we introduced a large scale real-world multi-person tracking dataset -- \texttt{PersonPath22}. The dataset is meticulously curated by (1) sourcing a set of videos that are diverse in terms of people density, camera angles, weather and scenery types as well as lighting conditions and (2) exhaustively annotating all persons in every frame with rigorous annotation and verification protocol that accommodates robust edge case handling. We demonstrated the value of the dataset by comparing it against existing datasets including MOT17~\cite{mot}, HiEve~\cite{hieve}, and MOT20~\cite{mot20}. Our dataset is a magnitude larger than the most popular MOT17 dataset in terms of unique person tracks, number of videos, and total video duration. We further performed in-depth analyses of existing state-of-the-art online trackers on our dataset and observed interesting cases where current online trackers fail to perform well. We hope that the publication of this dataset will spark a new wave of research towards developing more usable tracking models in real-world multi-person tracking. 

\paragraph{\textbf{Socially responsible usage of the dataset.}} \texttt{PersonPath22} dataset should primarily be used to improve person tracking algorithms, which can have a significant positive effect on many real-world video understanding problems including for example self-driving cars and human activity understanding. We ask the users of this dataset to use the data in a socially responsible manner, and request to not use the data to identify or generate biometric information of the people in the videos.

% \bing{Do we need to briefly mention the privacy and fairness issues of this dataset? It may be worth mentioning this.} \bing{I don't have a great idea where to start. If somebody has some idea, please write something here.}

% We've presented our tracking dataset that is an order of magnitude larger than the popular MOT17~\cite{mot}, HiEve~\cite{hieve}, and MOT20~\cite{mot20} datasets. This new dataset will enable the next generation of tracking algorithms.

\clearpage
\bibliographystyle{splncs04}
\bibliography{egbib}
\clearpage

\appendix

\noindent {\huge{Appendix}}

%%%%%%%%%%. SOURCING %%%%%%%%%%%%%%%%%%%
\section{Video Sourcing}

\subsection{Query keywords}
A pre-defined set of search keywords are used to query videos in stock video services (Fillerstock~\cite{fillerstock}, Pexels~\cite{pexels}, Pixabay~\cite{pixabay}). The complete list of keywords is as follows: 1), People walking in the street; 2), People walking in the mall; 3), People walking in the city; 4), People walking near market; 5), People walking inside mall; 6), People walking inside shops; 7), People crossing the street; 8), People walking in open space. We select this set specifically to query videos that include moving people rather than  static people for the interest of tracking applications.

\subsection{Videos Manual Selection}
This section adds details regarding the video manual selection process introduced in Sec.~\mainref{3} in the main paper. More in detail, our team of experts selected the videos to be part of the dataset by taking into account the following criterias:

\begin{itemize}

\item \textbf{Application Aligned.} We look for videos that appear to come from fixed connected home or city level cameras. We discard videos captured from television programs that contain advertisement interruptions, or videos with subtitles or that have been otherwise edited by software.
%We only kept videos that would be likely used in practice in real-world tracking applications. More in detail, we discarded web-sourced videos captured from television programs that contained advertisement interruptions, or videos with subtitles or other overlaid content. We also discarded videos that were heavily modified by video editing software.

\item \textbf{Moving crowds.} We try to strike a balance between (1) avoiding trivial videos with one or two people only, and (2) having over-crowded videos with hundreds of people, which we believe would have posed challenges at annotation time and lowered the ground truth accuracy. In particular, we discard videos with less than 5 people, and videos where more than half of the scene contained crowds, with less than 10\% of each person visible. We also favor the selection of videos containing moving people rather than static ones (sitting or standing), given the fact that the latter represents a trivial scenario for tracking.

\item \textbf{Occlusion.} We also favor the selection of videos containing high degrees of occlusions, where people become occluded for one or more seconds, and then reappear in the scene. This is done in order to promote the development of algorithms that can handle such situations. We include both partial and full occlusions, as well as person-to-person and person-to-object occlusions.

\item \textbf{Background variability.} Our selection process also ensures diversity in the background scene. For example, Virat~\cite{virat} contains 315 videos collected from the same 11 cameras at the same time of the day.
Videos from the same camera are highly redundant in both visual appearance as well as in motion and occlusion statistics. For these reasons, we select at most 3 videos per camera.
% VIRAT info: grep .mp4 /workplace/data/motion_efs/datasets/omni_data/virat/annotation/splits.json  |wc -l
% https://data.kitware.com/api/v1/file/56f581c88d777f753209c9ce/download

\item \textbf{Static vs moving cameras.} Our dataset focuses on the static-camera use-case, therefore we discard videos captured by moving cameras.

\item \textbf{Camera position.} We select videos from a large variety of camera positions and angles (from bird's-eye view to low-angle view).

\item \textbf{Environment conditions.} We select videos recorded at different times of the day (day/night) and different weather conditions (sunny/rain/snow/cloudy). In particular, we ensure a proper representation of videos captured at night or in foggy conditions which will challenge even the best state-of-the-art object detector.
\end{itemize}

\begin{table}[t]
    \centering
    \small
    \begin{tabular}{lcccccc}
        \toprule
        Data source & Number of videos  \\
        \midrule
        Fillerstock~\cite{fillerstock} & 101 \\
        Meva~\cite{meva} &	16 \\
        PathTrack~\cite{pathtrack}	& 26 \\
        Virat~\cite{virat}	& 9 \\
        Pexels~\cite{pexels}	& 78 \\
        Pixabay~\cite{pixabay}	& 6 \\
        \midrule
        Total & 236\\
        \bottomrule
    \end{tabular}
    
    \caption{Composition of our dataset: number of videos per data source}
    \label{tab:data_pools}
\end{table}

\subsection{Details on dataset composition}
Tab.~\ref{tab:data_pools} shows a detailed break-down of the number of videos per data source. Note that only~20\% of the videos were sourced from existing public datasets. In particular, we selected only 25 videos from Virat~\cite{virat} and Meva~\cite{meva} with at most 3 videos per unique camera.

\begin{figure}[t]
    \centering
    \includegraphics[width=1.0\textwidth]{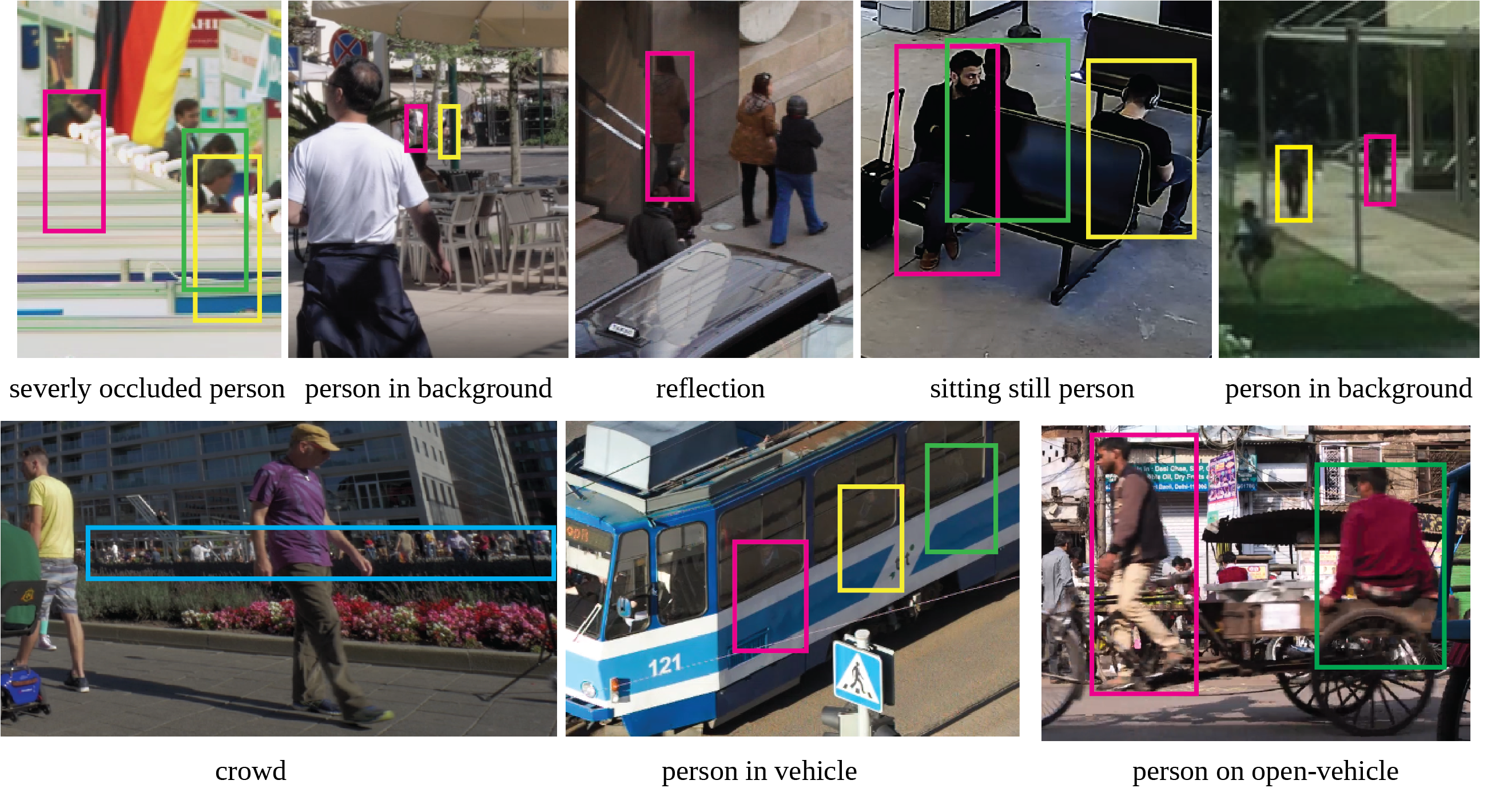}
    \caption{Visualization of person bounding boxes with special tags. Note that we only show the person boxes that have a special tag, which means that the persons that are not enclosed with a bounding box in the given examples are tagged with ``foreground person". }
    \label{fig:suppl_vis_tag}
\end{figure}

\section{Annotation Pipeline}

The definition of the track-level tag is as follows.
\begin{itemize}
    \item \textbf{Sitting/standing still person}. A person is sitting/standing without moving in the entire video, such as a person that is eating in a restaurant or is resting on the airport bench.
    \item \textbf{Person in vehicle}. A person is sitting in a moving or static vehicle and it is usually occluded.  Vehicles include buses, cars, trams and trains. 
    \item \textbf{Person on open-vehicle}. A person/baby is sitting on an open-vehicle whicn includes bicycles, motorbikes, rickshaw charts or stroller. They are usually fully visible.
    \item \textbf{Reflection}. A person is reflected in a surface such as a mirror or window.
    \item \textbf{Severely occluded person}. A person is severely occluded (less than 20\% visible) during the full duration of the video.
    \item \textbf{Person in background}. A person is  far away from camera and its identity is barely distinguishable without context. In this case, the person can appear within a \texttt{crowd} region, or it is severely occluded, or its size is too small.  
    % \joe{we need a better definition for this. Is this separate from crowds?}
    \item \textbf{Foreground person}. A moving person in the foreground which might get occluded from time to time and whose identity is recognizable most of the time. We primarily target tracking these person tracks. % \joe{do we even tag these specifically? I would just remove this as all other persons fall in the category} \bing{For completeness, I think we should include them.}
    % A person track that does not satisfy any of the above conditions. \joe{can we state this as a positive rather than negative? "A moving person in the foreground"?}\uta{+1}
\end{itemize}

In Fig.~\ref{fig:suppl_vis_tag}, we also show representative visual examples for the special tag.

\begin{figure}[t!]
\centering

\rotatebox[origin=c]{90}{\bfseries \scriptsize{PersonPath22 (train)}\strut}
    \begin{subfigure}{0.25\textwidth}
        \centering
        \includegraphics[width=1.1\textwidth]{Figs/dataset_stat/amazon_track_speed.png}
    \end{subfigure}%
    \begin{subfigure}{0.25\textwidth}
        \centering
        \includegraphics[width=1.1\textwidth]{Figs/dataset_stat/amazon_track_occlusion.png}
    \end{subfigure}%
    \begin{subfigure}{0.25\textwidth}
        \centering
        \includegraphics[width=1.1\textwidth]{Figs/dataset_stat/amazon_track_duration.png}
    \end{subfigure}
    
\vspace{-1.2em}
\rotatebox[origin=c]{90}{\bfseries \scriptsize{MOT17} \strut}
    \begin{subfigure}{0.25\textwidth}
        \centering
        \includegraphics[width=1.1\textwidth]{Figs/dataset_stat/mot17_speed.png}
    \end{subfigure}%
    \begin{subfigure}{0.25\textwidth}
        \centering
        \includegraphics[width=1.1\textwidth]{Figs/dataset_stat/mot17_occlusion.png}
    \end{subfigure}%
    \begin{subfigure}{0.25\textwidth}
         \centering
         \includegraphics[width=1.1\textwidth]{Figs/dataset_stat/mot17_duration.png}
    \end{subfigure}

\vspace{-1.2em}
\rotatebox[origin=c]{90}{\bfseries \scriptsize{MOT20} \strut}
    \begin{subfigure}{0.25\textwidth}
        \centering
        \includegraphics[width=1.1\textwidth]{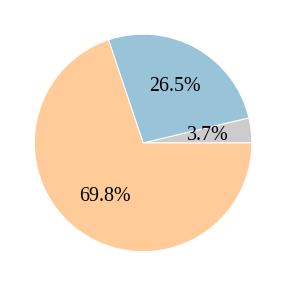}
    \end{subfigure}%
    \begin{subfigure}{0.25\textwidth}
        \centering
        \includegraphics[width=1.1\textwidth]{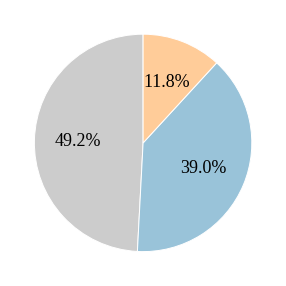}
    \end{subfigure}%
    \begin{subfigure}{0.25\textwidth}
         \centering
         \includegraphics[width=1.1\textwidth]{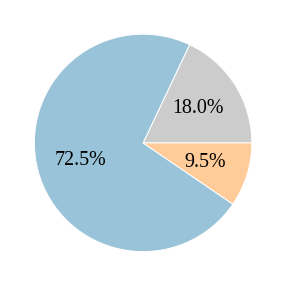}
    \end{subfigure}
    
    \vspace{-1.2em}
\rotatebox[origin=c]{90}{\bfseries \scriptsize{HiEve} \strut}
    \begin{subfigure}{0.25\textwidth}
        \centering
        \includegraphics[width=1.1\textwidth]{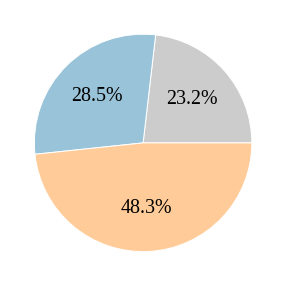}
    \end{subfigure}%
    \begin{subfigure}{0.25\textwidth}
        \centering
        N/A
    \end{subfigure}%
    \begin{subfigure}{0.25\textwidth}
         \centering
         \includegraphics[width=1.1\textwidth]{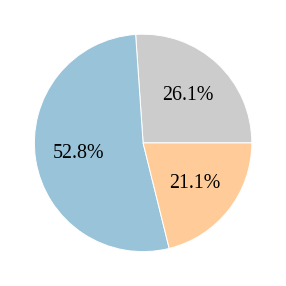}
    \end{subfigure}
    
    \vspace{-1.2em}
\rotatebox[origin=c]{90}{\bfseries \scriptsize{PersonPath22(test)} \strut}
    \begin{subfigure}{0.25\textwidth}
        \centering
        \includegraphics[width=1.1\textwidth]{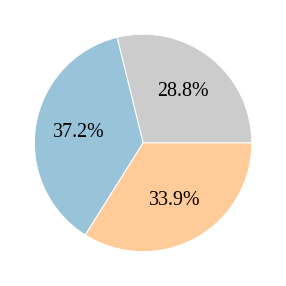}
        \caption{\scriptsize Track speed}
        \label{fig:suppl_dataset_stat_speed}
    \end{subfigure}%
    \begin{subfigure}{0.25\textwidth}
        \centering
        \includegraphics[width=1.1\textwidth]{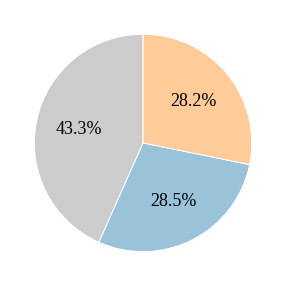}
        \caption{\scriptsize Occlusion duration}
        \label{fig:suppl_dataset_stat_occlusion}
    \end{subfigure}%
    \begin{subfigure}{0.25\textwidth}
         \centering
         \includegraphics[width=1.1\textwidth]{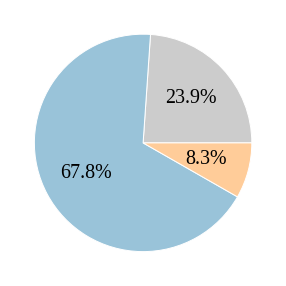}
        \caption{\scriptsize Track duration}
        \label{fig:suppl_dataset_stat_length}
    \end{subfigure}
    
    \caption{\small Comparison of track-level statistics of person tracks among different datasets.  Note that we are unable to estimate the occlusion duration on HiEve dataset~\cite{hieve} as meta data related to occlusion of each person box is not available.
    % \uta{why not also adding MOT17 here? I know we already have it in the main paper, but for completeness I would add all since we have the space + (in case that wasn't planned yet) it would be good to also explain why we don't have any occlusion information from HiEve}
    \label{fig:suppl_track_stat}
    }
\end{figure}

\section{Dataset}

\begin{table}[t]
    \centering
    \begin{tabular}{lccccc}
    \toprule
         \multirow{2}{*}{Dataset} & \multirow{2}{*}{\#Videos} & Length & \#Annotated  & \#Person & Min \\
         & & (secs) & Frames  & Tracks & Res.  \\
         \midrule
         PersonPath22 (All) & 236 & 8,334 &  200,769  & 12,150 & 720x480  \\
         PersonPath22 (Train) & 138 & 4,736 &  118,685  & 7,096 & 720x480  \\
         PersonPath22 (Test) & 98 & 3,598 &  82,084  & 5,054 & 720x480  \\
    \bottomrule
    \end{tabular}
    \caption{\small Key statistics of PersonPath22 dataset. Annotated Frames refer to the frames that are manually annotated and those
that are automatically interpolated and then manually verified.}
    \label{tab:train_test_stat}
    
\end{table}

\subsection{Statistics of train/test split}
In Tab.~\ref{tab:train_test_stat}, we show the key statistics of the full PersonPath22 dataset as well as its train and test splits.  In Fig.~\ref{fig:suppl_track_stat}, we extend the Fig.~\mainref{4} in the main paper to show the \mbox{track-level} statistics for our dataset (train / test split), MOT17~\cite{mot}, MOT20~\cite{mot20} and HiEve~\cite{hieve}. Note that we are unable to estimate the occlusion duration on HiEve dataset as meta data related to occlusion is not provided. In both MOT17~\cite{mot} and MOT20~\cite{mot20}, we define a person as fully occluded when the provided \texttt{visibility} score (between 0 and 1) is below $0.05$.

Overall, our dataset includes higher-proportion of medium-to-long occlusion cases, which we think are of significant interest in real-world person tracking scenarios.

% \subsection{Video demos }
% We attach a few representative videos by overlaying their manually annotated person tracks. Note that the unique identifier of each person is visualized with a unique color in each video, and the black-coded person tracks are not used during model training and evaluation. In addition, the person boxes are not shown when they are fully occluded.
% \bing{We mentioned in the main paper that only boxes with ``foreground person" tag are used for model evaluation, but we also include ``sitting" person during evaluation. So probably we should avoid showing sitting person in the demo videos.}

%%%%%%%%%%. TRAINING %%%%%%%%%%%%%%%%%%%
\section{Experiments}

\subsection{Ignore person tracks}
During training, we ignore person tracks that are tagged with `\texttt{person in vehicle}', `\texttt{severely occluded person}' and `\texttt{person in background}', as the corresponding person is barely recognizable. During evaluation, on top of those person tracks, we further ignore person tracks that are tagged with `\texttt{reflection}' and `\texttt{person on open-vehicle}' as we find those person tracks are less interesting in real-world applications. In addition, we also ignore person bounding boxes that are included in a `\texttt{crowd}' region during evaluation.   

\subsection{Implementation details}

\paragraph{CenterTrack~\cite{centertrack}.} We use the official model made available by the authors, which was trained on the CrowdHuman dataset~\cite{shao2018crowdhuman} using an input resolution of $512 \times 512$. We then fine-tune the model on PersonPath22 dataset using a learning rate of $1.25e^{-4}$ for 10 epochs using mini-batch size 32, increasing the input resolution to $960 \times 544$. We set $\lambda_{fp}=0.1$, $\lambda_{fn}=0.4$ at training time, and $\theta = 0.5$ and $\tau = 0.5$ at evaluation time, and enabled both the input tracking heatmap as well as the amodel box training and inference, which was found to improve the tracking metrics by the original authors.

\paragraph{SiamMOT~\cite{siammot}}. We start with the official CrowdHuman-pretrained model made available by the authors, and we fine-tune it on PersonPath22 dataset by using a learning rate of $0.01$ for $10K$ iterations with SGD with momentum. The learning rate is decreased by a factor of $10$ after $6K$ and $8K$ iterations, respectively. All training details remain the default as suggested by the paper~\cite{siammot}. During inference, we set the linking confidence $\alpha = 0.4$, the detection confidence $\beta = 0.6$ and keep a trajectory active until it is unseen for $\tau = 30$ frames. We resize the image such that its shorter side has $800$ pixels while its longer side does not exceed $1,500$ pixels.

\paragraph{FairMOT~\cite{fairmot}.} We finetune the official CrowdHuman-pretrained model on PersonPath22 dataset using a learning rate of $0.001$ for $20$ epochs with a batch size of $32$. The learning rate is decreased by a factor of $10$ after $15$ epochs. All other training and inference details remain the same as in the official FairMOT github repository.

\paragraph{IDFree~\cite{pointid}.} We train the model only with CrowdHuman dataset, in which the person embedding is trained with the proposed dense contrastive loss. We train the model with $25K$ iterations with a batch size as $16$ and initial learning rate $0.01$. The learning rate is annealed by $0.1$ after $17.5K$ and $22.5K$ iterations.  During inference, we set the linking confidence $\alpha = 0.45$, the detection confidence $\beta = 0.6$ and keep a trajectory active until it is unseen for $\tau = 30$ frames. We resize the image such that its shorter side has $900$ pixels while its longer side does not exceed $1,500$ pixels.

\paragraph{TrackFormer~\cite{trackformer}.} We first pre-train the trackformer with the CrowdHuman dataset for 80 epochs. Furthermore, we fine-tune the model on the joint CrowdHuman and PersonPath22 dataset for another 40 epochs. We start the training / pre-training with the learning rate $0.0002$ and anneal it by $0.1$ after $50$ during model pre-training and $10$ epochs during model fine-tuning respectively. During inference, we resize the image such that its shorter side has $800$ pixels while its longer side does not exceed $1,333$ pixels. All other training and inference details remain the same as in the official TrackFormer github repository.

\paragraph{ByteTrack~\cite{bytetrack}.} For ByteTrack, the detector is YOLOX~\cite{yolox} with YOLOX-X as the backbone and COCO-pretrained model as the initialized weights. We train YOLOX with 80 epochs on PersonPath22 training set. The optimizer is
SGD with weight decay of $5 \times 10^{-4}$ and momentum of $0.9$. The initial learning rate is $0.001$ with 1 epoch warm-up and cosine annealing schedule. All other training and inference details remain the same as in the official ByteTrack github repository. 
\end{document}